\documentclass[conference]{IEEEtran}
\IEEEoverridecommandlockouts
\usepackage{cite}
\usepackage{amsmath,amssymb,amsfonts}
\usepackage{algorithmic}
\usepackage{graphicx}
\usepackage{textcomp}
\usepackage{xcolor}
\usepackage{algorithm}
\def\BibTeX{{\rm B\kern-.05em{\sc i\kern-.025em b}\kern-.08em
    T\kern-.1667em\lower.7ex\hbox{E}\kern-.125emX}}
    
\usepackage{booktabs}
\usepackage{amsmath}
\usepackage{amsfonts}
\usepackage{bbding}
\usepackage{makecell}
\usepackage{multirow}
\usepackage{newfloat}
\usepackage{listings}
\begin{document}

\title{Streaming Video Temporal Action Segmentation \\ In Real Time}

\author{\IEEEauthorblockN{1\textsuperscript{st} Wujun Wen}
\IEEEauthorblockA{
\textit{Dalian University of Technology}\\
Dalian, China\\
thinksky@mail.dlut.edu.cn}
\and
\IEEEauthorblockN{2\textsuperscript{nd} Yunheng Li}
\IEEEauthorblockA{
\textit{Dalian University of Technology}\\
Dalian, China\\
liyunheng@mail.dlut.edu.cn}
\and
\IEEEauthorblockN{3\textsuperscript{rd} Zhuben Dong}
\IEEEauthorblockA{
\textit{Dalian University of Technology}\\
Dalian, China\\
201762014@mail.dlut.edu.cn}
\and
\IEEEauthorblockN{4\textsuperscript{th} Lin Feng}
\IEEEauthorblockA{
\textit{Dalian University of Technology}\\
Dalian, China\\
fenglin@dlut.edu.cn}
\and
\IEEEauthorblockN{5\textsuperscript{th} Wanxiao Yang}
\IEEEauthorblockA{
\textit{Dalian University of Technology}\\
Dalian, China\\
yang224425@mail.dlut.edu.cn}
\and
\IEEEauthorblockN{6\textsuperscript{th} Shenlan Liu}
\IEEEauthorblockA{
\textit{Dalian University of Technology}\\
Dalian, China\\
liusl@dlut.edu.cn}
}

\maketitle

\begin{abstract}
Temporal action segmentation (TAS) is a critical step toward long-term video understanding. Recent studies based on the pre-extracted video features instead of raw video image information. However, those models are trained complicatedly and limit application scenarios. It is hard for them to segment human actions of video in real time because they must work after the full video features are extracted. As the real-time action segmentation task is different from TAS task, we define it as streaming video real-time temporal action segmentation (SVTAS) task. In this paper, we propose a real-time end-to-end multi-modality model for SVTAS task. More specifically, we incrementally propose three frameworks to solve the SVTAS task and enhance the model performance step-by-step. To the best of our knowledge, it is the first multi-modality real-time temporal action segmentation model. Under the same evaluation criteria as full video TAS, our model segments human action in real time with less than 40\% of state-of-the-art model computation and achieves 90\% of the accuracy of the full video state-of-the-art model. Code is available at https://github.com/Thinksky5124/SVTAS.git.
\end{abstract}


\section{Introduction}

Temporal action segmentation (TAS) is a task of video understanding, which plays a vital role in security monitoring, sports analysis, and so on. Its goal is to classify every frame of video.

\begin{figure}[t]
    \centering
    \includegraphics[width=\columnwidth]{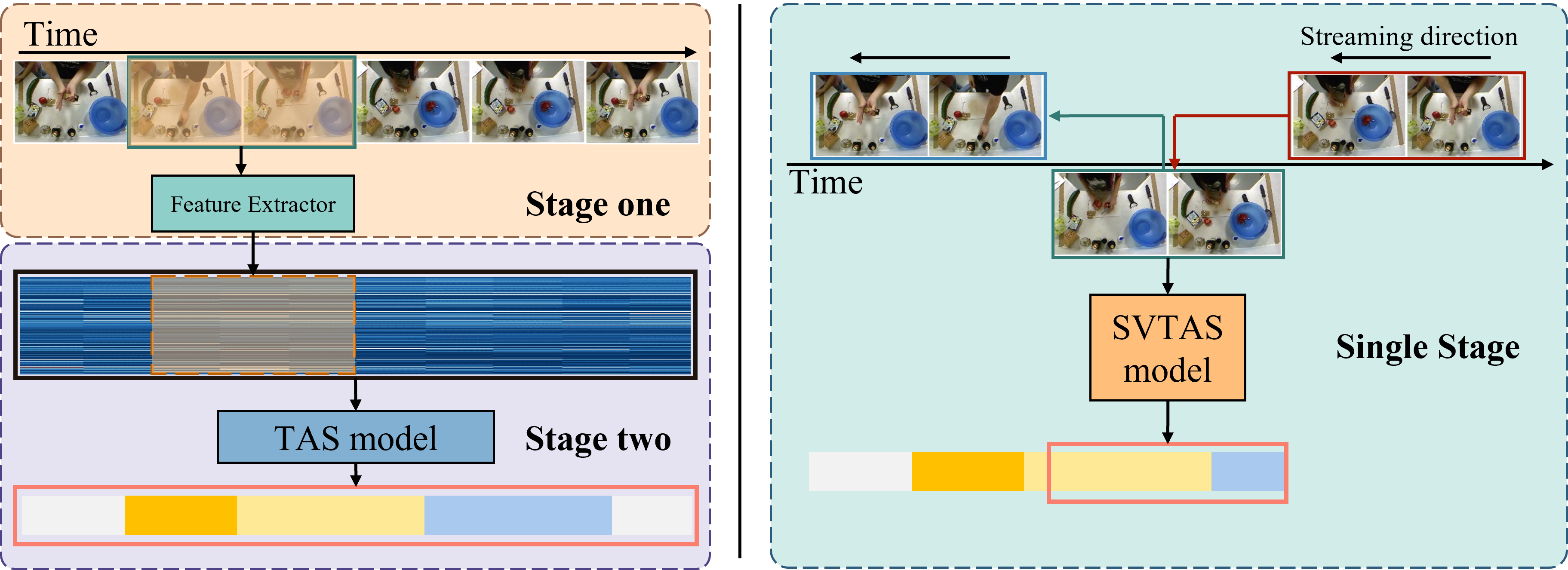}
    \caption{Two stage model vs. single stage model. The left paradigm could only be applied for a full video, however, the right paradigm could be applied for real-time or streaming video besides full video.}
    \label{fig1}
\end{figure}

Despite the great success of TAS techniques for full video sequences in recent years, these models are trained in two stages and only used on full video sequences \cite{farha2019ms,li2022bridge,yi2021asformer,ahn2021refining}. The two-stage training process first extracts frame-level features by training an action recognition model which requires manual trimming and annotating for extra recognition training set. Then, a TAS model generates the frame-level action labels. This process can only be applied to offline scenes, but limits streaming or real-time scenarios. Recent studies have used pre-trained models as feature extractors, which would be free of additional manual video trimming. However, it is still not applicable to streaming or real-time segmentation scenarios and performs worse than the models after fine-tuning in the extra recognition training set. Therefore, we propose an alternative scheme, SVTAS for TAS task. As represented in Figure \ref{fig1}, the two-stage approaches have a complex pipeline process and can only be executed serially, which cannot be applied to real-time scenarios. In contrast, the single-stage approaches use an end-to-end model and can be applied to a variety of scenarios. However, the single-stage approaches can only rely on current information while the two-stage approaches can access to the global time-serious relationship, which makes the single-stage approaches more challenging.

The model under the definition of SVTAS can be applied in offline, online, streaming, and real-time scenarios, etc. And, an end-to-end training model can avoid complicated training steps. To be more specific, the advantages of end-to-end training SVTAS models include a simple and efficient training process, a significant reduction in network latency (time from model input to model output), and extensive application in most video scenarios. But the end-to-end training SVTAS task is extremely more challenging than existing related tasks. Firstly, since only the raw information of a small segment is available, the model will not get the global time-serious relationship when segments a full video human action. It will significantly weaken the model ability to classify long-term action and infer sequential action. Both issues will significantly reduce model accuracy in our experimental results. Secondly, the over-segmentation problem \cite{farha2019ms} appears because of the narrow segmented window. It will make the model segment complete action with unbearable burrs. Finally, the naive scheme combining action recognition model with TAS model is powerless for SVTAS task, which fails to classify frame-level action precisely (See Table \ref{tab1}). Although two-stage SVTAS can achieve satisfactory performance, an independent feature extractor is inevitable in complex training steps.

This paper proposes a generalizable end-to-end training framework by selective feature extractors and temporal model for SVTAS task. And to response the challenge of losing global time-serious relationship in SVTAS task, we use a temporal convolution network with memory cache (memory tcn) to store short-time temporal information. Then, inspired by \cite{radford2021learning,li2022bridge}, we explore a new multi-modality way to model long-time past information. By mixing natural language and image modality, our multi-modality real-time model, Transeger, obtains better results than any other models in SVTAS task.

In summary, our contributions are as follows:
\begin{itemize}
\item \textbf{Streaming video TAS in real time:}
Unlike full-sequence TAS approaches, we proposed a framework to realize real-time TAS technique for streaming videos. Our framework changes the growth of network latency and space complexity of the TAS model from linear $O(n)$ to constant $O(1)$ by setting the chunk size of the model, which significantly reduces the network latency and broadens the application scenarios of TAS.

\item \textbf{End-to-end train:}
Our framework can be trained end-to-end, which simplifies the training processes. We complete the training of the model using only one stage, without adding additional manual editing, annotation, and optical flow extraction work. And, our framework even reduces the number of model parameters and computation.

\item \textbf{Multi-modality:}
The multi-modality SVTAS model we propose, Transeger, mitigates the problem that model loses global time-series relationship by importing natural language modality to model past information. To the best of our knowledge, it is the first multi-modality real-time temporal action segmentation model.
\end{itemize}

\section{Related Work}
\subsection{Action Recognition}
Action Recognition (AR) aims to give a single action label to a trimmed human action video \cite{sun2022human}, so it is a classification task in which all frames of the video correspond to a label (many-to-one). \cite{simonyan2014two} uses convolution neural networks to fuse RGB modality and optical flow modality in the video to recognize human actions. \cite{kondratyuk2021movinets} proposes a stream method to recognize human actions by fusing segments of the video, with the purpose of reducing the space complexity of the action recognition model without declining the accuracy. The action recognition models follow the many-to-one paradigm, while SVTAS task requires the model to be a one-to-one paradigm. The experiments in Table \ref{tab1} will show that directly transferring action recognition model to SVTAS task performs poorly.

\subsection{Temporal Action Segmentation}
Temporal Action Segmentation (TAS) aims to classify each frame from the entire sequence of an untrimmed video, so it is a classification task (one-to-one) where each frame in the video corresponds to a relevant label. \cite{farha2019ms} uses a multi-stage temporal convolution network to model full-sequence video features, which greatly improves the performance of TAS. \cite{li2022bridge} uses contrastive learning between natural language modality and image modality to train the feature extractor so that the full-sequence video features extracted by the feature extractor are easier for the TAS model to segment action. Although the above models could achieve desirable performance, they all follow the paradigm of two-stage training and a full sequence of video segmentation, which could not meet the requirements of real-time scenarios.

\subsection{Online Action Detection}
Online Action Detection (OAD) aims to detect whether an untrimmed human action video contains a target action and gives the category of the action \cite{hu2022online}, so it is a classification task of all frames of video corresponding to one label (many-to-one). \cite{wang2021oadtr} combines action prediction with online action detection, and applies Transformer architecture to improve model performance. Although online action detection is somewhat similar to SVTAS, the experiments in Table \ref{tab1} will show that direct transferring online action detection model is also under-performing.

\section{Method}
In this section, we first give the definition of SVTAS task, then describe our proposed end-to-end training framework, and finally depict the multi-modality real-time TAS model for streaming video (Transeger).

\subsection{Task Definition}
SVTAS task refers to classify each action frame for a given streaming image sequence $ X_{j} = [x_{i}, x_{i+1}, \cdots, x_{i+k}]$ in chronological order, where $x_{i}$ represents $i^{th}$ frame of the video. $X_j$ is the $j^{th}$ non-overlapping segment in the video $\{x_{i}|i=0,1,2,\cdots,T - 1\}$ and contains $k$ frames, where $T$ represents the length of video, $j\in \{0, 1, 2, \cdots, \lceil \frac{T}{k} \rceil\}\}$. The result of the task is an action label sequence corresponding to each frame $x_{i}$, noted as $ L_{j} = [l_{i}, l_{i+1}, \cdots, l_{i+k}]$, where $l_{i}$ represents the action label of the $i^{th}$ frame in the video. And all labels of the video are donated as $L$. If the model is defined as $\mathcal{M}(\cdot)$, the mathematical expression of the task is $L = \{\cup_{j=0}^{\lceil \frac{T}{k} \rceil} L_{j}|L_{j}=\mathcal{M}(X_{j})\}$.

\subsection{Streaming End-to-End Train Framework (SETE)}

\begin{figure}[hbp]
    \centering
    \includegraphics[width=\columnwidth]{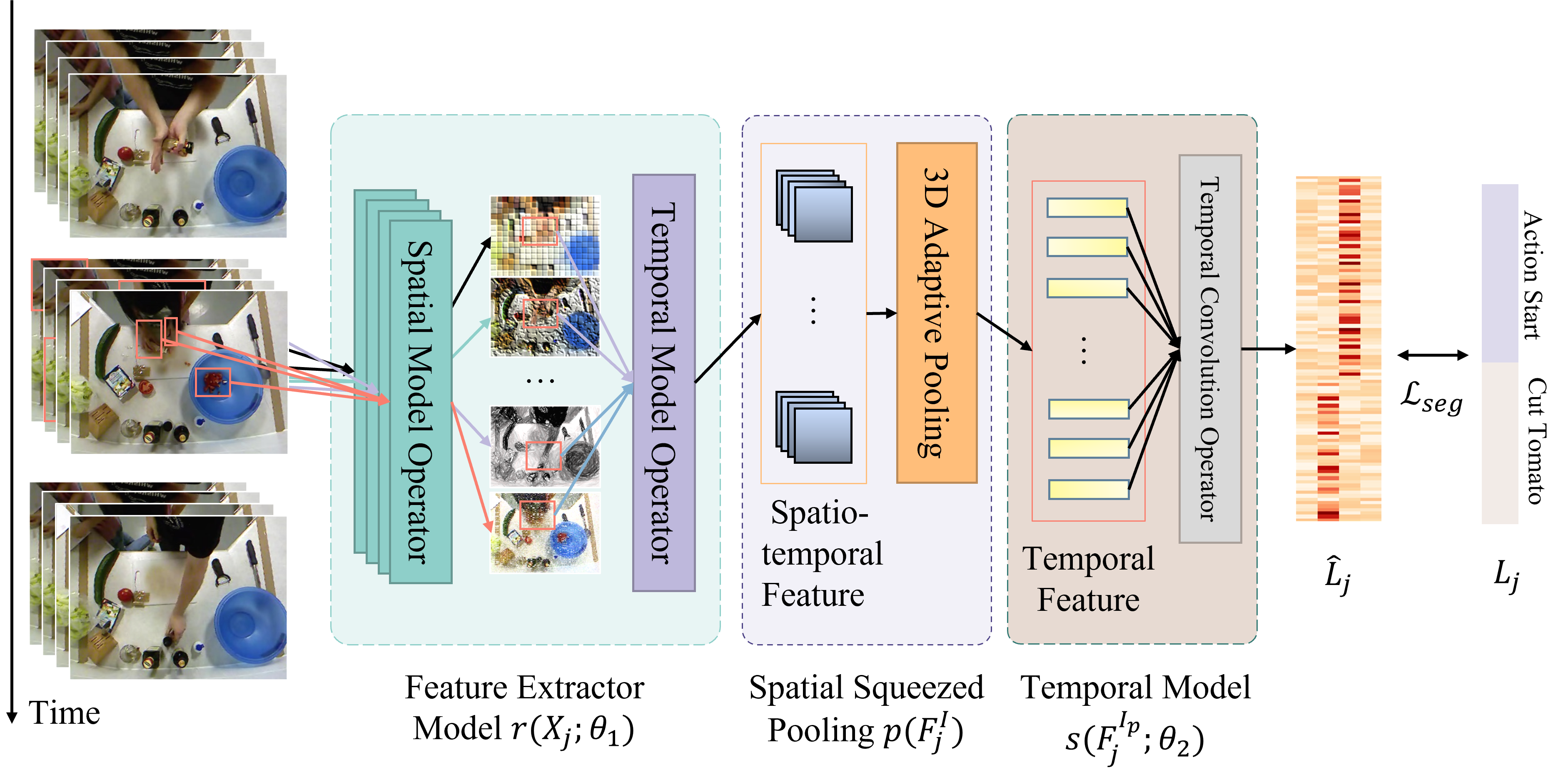}
    \caption{Streaming end-to-end training framework. The output is a matrix that is made up of the probability vector of each frame.}
    \label{fig2}
\end{figure}

This section describes the single stage end-to-end training framework we have constructed, which consists of a feature extractor model, spatial squeezed pooling, and a temporal model. See Figure \ref{fig2}.

\subsubsection{Feature Extractor Model}
Although feature extractor models that can only extract spatial features (e.g., ResNet \cite{he2016deep}, ViT \cite{ dosovitskiy2020image}, etc.) perform equally to models that have both spatial and temporal modeling capability (e.g., I3D \cite{carreira2017quo}, TSM \cite{lin2019tsm}, etc.) when segment action individually, we claim that as a feature extractor, a model with the ability to model space and time is required for SVTAS task (details can be referred to Table \ref{tab1}). Models with both spatial and temporal modeling capability generally employ 3D convolution network which models space and time simultaneously \cite{tran2018closer}. We claim that for the SVTAS task, the feature extractor models which could model time and space separately will perform better in the SVTAS tasks, e.g., TSM \cite{lin2019tsm}, TimeSformer \cite{bertasius2021space}, etc. We think it is because the SVTAS task is a frame-level classification task, which needs to model not only the similarity between frames but also the difference. Mathematically express the feature extractor model in the following form $F_j^I = r(X_j; \theta_1)$, where $F_j^I=[I_{i},I_{i+1} \cdots, I_{i+k}]$ represents the image features of the $j^{th}$ segment of the streaming video, $r(\cdot; \theta_1)$ denotes the feature extractor, and $X_j$ indicates the image of the $j^{th}$ segment of the streaming video.

\subsubsection{Spatial Squeezed Pooling}
Spatial squeezed pooling aims to pool the space information globally and averagely, leaving only temporal information. We claim that for SVTAS task, spatial information is unnecessary for final time-series segmentation because the difference between frames has been modeled by the feature extractor. Spatial squeezed pooling reduces the video memory footprint of graphics processing unit (VRAM) while maintaining the performance of model (See section 4.4). We define the mathematical expression for spatially squeezed pooling as follows: $F_j^{I_p} = p(F_j^I)$, where $p(\cdot)$ denoting spatial squeezed pooling and $F_j^{I_p}$ denoting the temporal features of the $j^{th}$ segment of the streaming video.

\subsubsection{Temporal Model}
Various temporal models can be applied for SVTAS task, for example, temporal convolution networks, long short-term memory networks, etc. In this paper, unlike TAS task, a single-stage temporal convolution network is selected for SVTAS task. The TAS task often applies a multi-stage technique to refine the results of the network to improve the model's discrimination of action completeness. But directly applying the multi-stage technique to SVTAS task not only fails to improve the discrimination of action completeness score but also significantly degrades the model performance in all aspects (See Table \ref{tab1}). We think as the segmentation windows become small, there is no need multi-stage technique to refine results. In addition, to meet the challenge that SVTAS task can only obtain raw video information at the current moment, we select a temporal convolution network with memory cache. Specifically, the temporal convolution operator typically uses padding techniques when the length of the time series needs to be maintained. We turn the usual case, zero padding, to padding the previous streaming video segment features at the beginning of the current streaming video temporal features. With the memory cache that stores short-time information of the previous video segment, temporal model could segment human action better. The temporal model can be expressed as follows $\widehat{L}_{j} = s(F_j^{I_p}; \theta_2)$, where $s(\cdot; \theta_2)$ represents the temporal model and $\widehat{L}_{j}$ denotes the model inference result of the $j^{th}$ segment of the streaming video.

\begin{figure}[htb]
    \centering
    \includegraphics[width=\columnwidth]{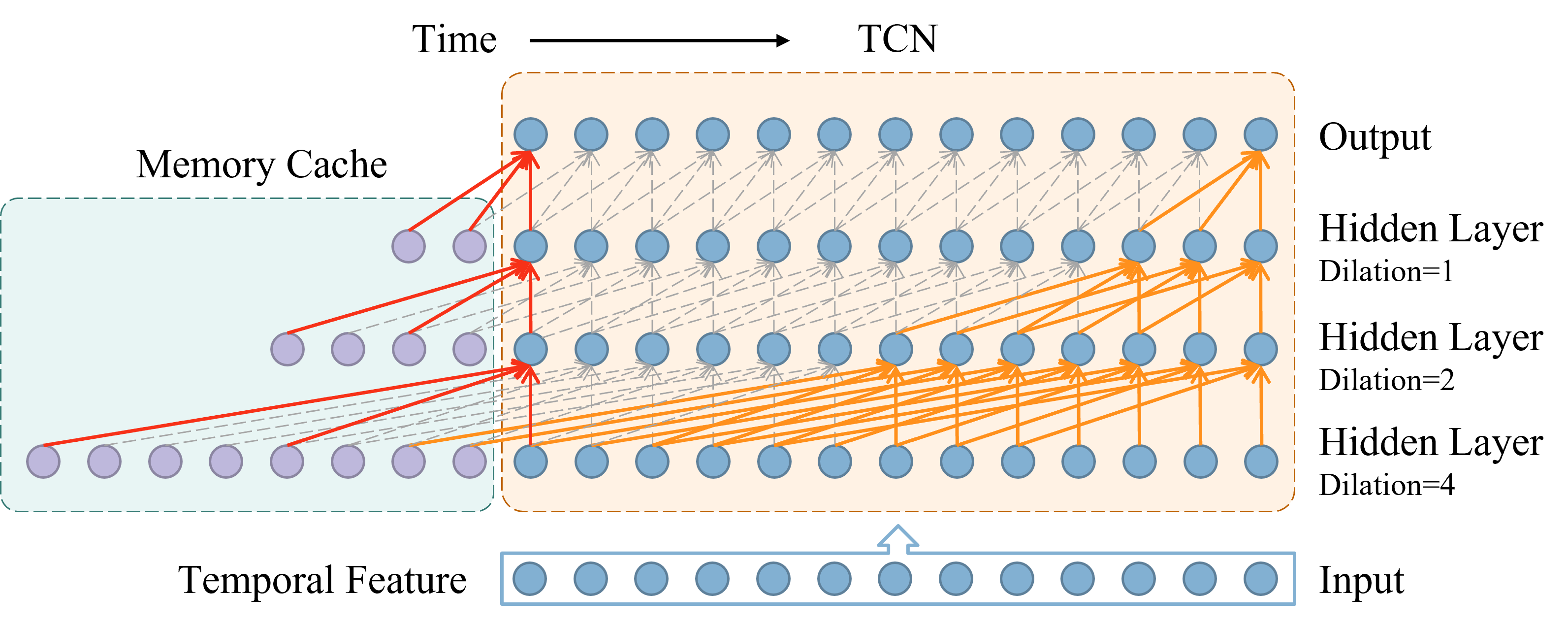}
    \caption{Temporal convolution network with memory cache. The orange line shows this information of frames is aggregated into the latest frame.}
    \label{fig3}
\end{figure}

\subsubsection{Loss Function}
The loss function \cite{farha2019ms} is divided by $\lceil \frac{T}{k} \rceil$ to make it work for streaming video end-to-end training. We define it as $\mathcal{L}_{seg}(L_{j}, \widehat{L}_{j})$.

\subsection{Multi-modality End-to-End Train Framework (METE)}

\begin{figure}[htpb]
    \centering
    \includegraphics[width=\linewidth]{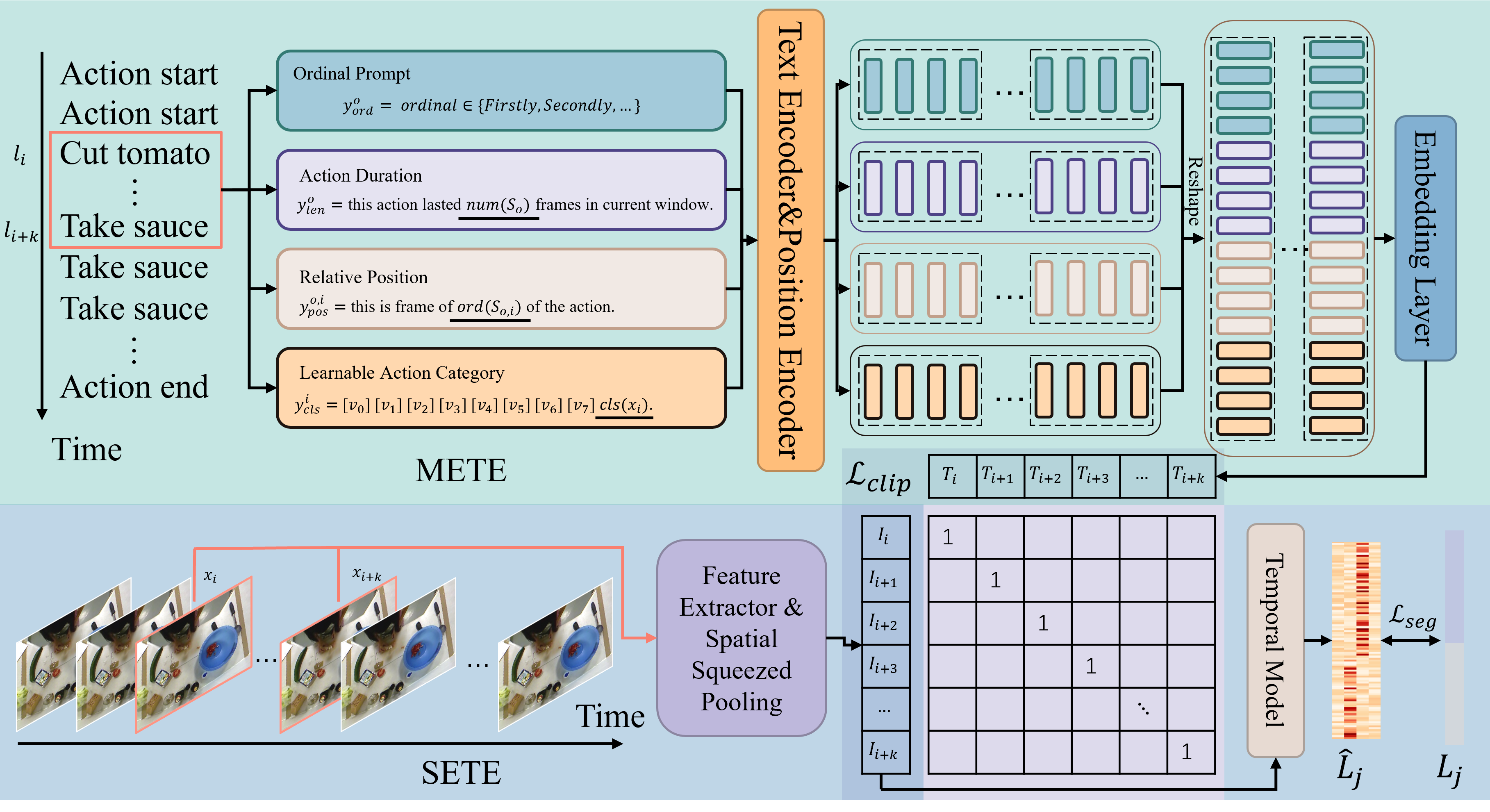}
    \caption{Multi-modality end-to-end train framework.}
    \label{fig4}
\end{figure}

Based on the SETE, we propose a training framework combining natural language modality and image modality for SVTAS task, as shown in Figure \ref{fig4}. This exposition part will serve as a paving part for the Transeger model.

The METE consists of an end-to-end SVTAS model, a mix prompt engineering, and a text encoder. The first is the end-to-end SVTAS model, which is the aforementioned SETE so it will not be repeatedly described. The second is the mix prompt engineering, which consists of a mixture of hand-designed templates and learnable word embedding vectors. The last is the text encoder. With the success of Transformer in natural language processing, this paper uses the Transformer structure as a text encoder.

\subsubsection{Mix Prompt Engineering}
Prompt engineering refers to the design of an input text template that embeds the expected output strings as fill-in-the-blank formats (e.g., cut tomato). Inspired by \cite{zhou2021learning}, we improve prompt engineering proposed by \cite{li2022bridge} which fixes template of sequential information. Our prompt engineering firstly introduces action segment information, then adds the length information of action segment and relative position information of frames, and finally turns the information of action categories into a learnable form. We call it mix prompt engineering because there are both fixed and learnable parts in the template.

\begin{itemize}
\item \textbf{Ordinal Prompt}
captures the position of each action segment in the current video sequence. We use the set of ordinal words \textit{\{Firstly, Secondly, ... \}} as a template, noting it as $y_{ord}^{o}$, where $o$ indicates that the current action is the $o^{th}$ action in the current video sequence. For an action segment sequence, we mark the sequence ordinal prompt as $y_{ord}^o\in \{y_{ord}^{1},y_{ord}^{2},y_{ord}^{3},\cdots\}$, where $o = ord(x_{i})$ denotes taking the sequence number of the action segment from the $i^{th}$ frame in the current video sequence.

\item \textbf{Action Duration Prompt}
captures the length of the current action segment. This information helps the model infer the category of the action by the duration of the action. We use \textit{this action lasted \underline{\{$num(S_{o})$\}} frames in current window} as a template, labeling it as $y_{len}^{o}$, where $S_{o}$ denotes the set of frames of the $o^{th}$ action, $S_{ o}=\{x_{o_{s}}, x_{o_{s}+1}, \cdots,x_{o_{e}}\}$, $o_{s}$ denotes the start frame of the $o^{th}$ action in the current video sequence, and $o_{e}$ denotes the end frame of the $o^{th}$ action in the current video sequence.

\item \textbf{Relative Position Prompt}
captures the relative position of the current frame in the current action segment. This information helps the model to classify some reversible actions, e.g., opening and closing a bottle cap. Reversible actions, which are mirrored over time, are easily confused the model when models are  inferring. After adding the relative position prompt, the model can distinguish the mirrored actions by the ordinal relationship between the front and back frames. We use \textit{this is frame \underline{\{$ord(S_{o,i})$\}} of the action} as a template, labeling it as $y_{pos}^{o,i}$, where $ord(S_{o,i})$ indicates taking the ordinal number for the $i^{th}$ frame of the $o^{th}$ action.

\item \textbf{Learnable Action Category Prompt}
captures the category information of the current action segment. Following the guidelines of \cite{zhou2021learning}, we use $\chi_0 \, \chi_1 \, \chi_2 \, \chi_3 \, \chi_4 \, \chi_5 \, \chi_6 \, \chi_7 \, $\textit{\underline{\{$cls(x_{i} )\}$}} as the template, which is labeled as $y_{cls}^{i}$, where $cls(x_{i})$ denotes the action category that takes from the $i^{th}$ frame, $\chi_0,\chi_1,\chi_2,\cdots,\chi_7$ denotes the learnable word embedding vectors, and eight learnable word vectors are used in this paper.

\end{itemize}

We connect the above prompts all together in the direction of the natural order of sentence. The combinations are varied obviously. We use follow order as a practice $pr_{i} = [y_{ord}^{o}, y_{len}^{o}, y_{pos}^{o,i}, y_{cls}^{i}]$ and $Pr_{j} = [pr_{i},pr_{i+1}, \cdots, pr_{i+k}]$.

\subsubsection{Text Encoder}
The text encoder converts the text string generated by the prompt engineering algorithm into text features $F_{j}^{T}=[T_{i},T_{i+1}, \cdots, T_{i+k}]$, and we follow the way in \cite{radford2021learning} for text embedding and text feature extraction. Transformer is used as text encoder and we set encoder dimension 512. We use the conclusion in \cite{zhou2021learning} that using the suffix form to fill in the category labels is better than others. We define it as $F_{j}^{T} = t(Pr_{j}; \theta_{3})$, where $t(\cdot; \theta_3)$ represents text encoder.

\subsubsection{Loss Function}
We use the loss function of the contrast between text features and image features described in \cite{radford2021learning} and define it as $\mathcal{L}_{clip}(F_{j}^{Img}, F_{j}^{Txt})$, where $F_{j}^{Img}$ is image features and $F_{j}^{Txt}$ is text features. When model is trained, the model will be supervised by both loss function, $\mathcal{L}_{clip}$ and $\mathcal{L}_{seg}$.

Although METE introduces contrast supervision which makes the accuracy of the model decrease, it provides suggestions for multi-modality real-time TAS.

\begin{figure*}[t]
\centering
\includegraphics[width=\linewidth]{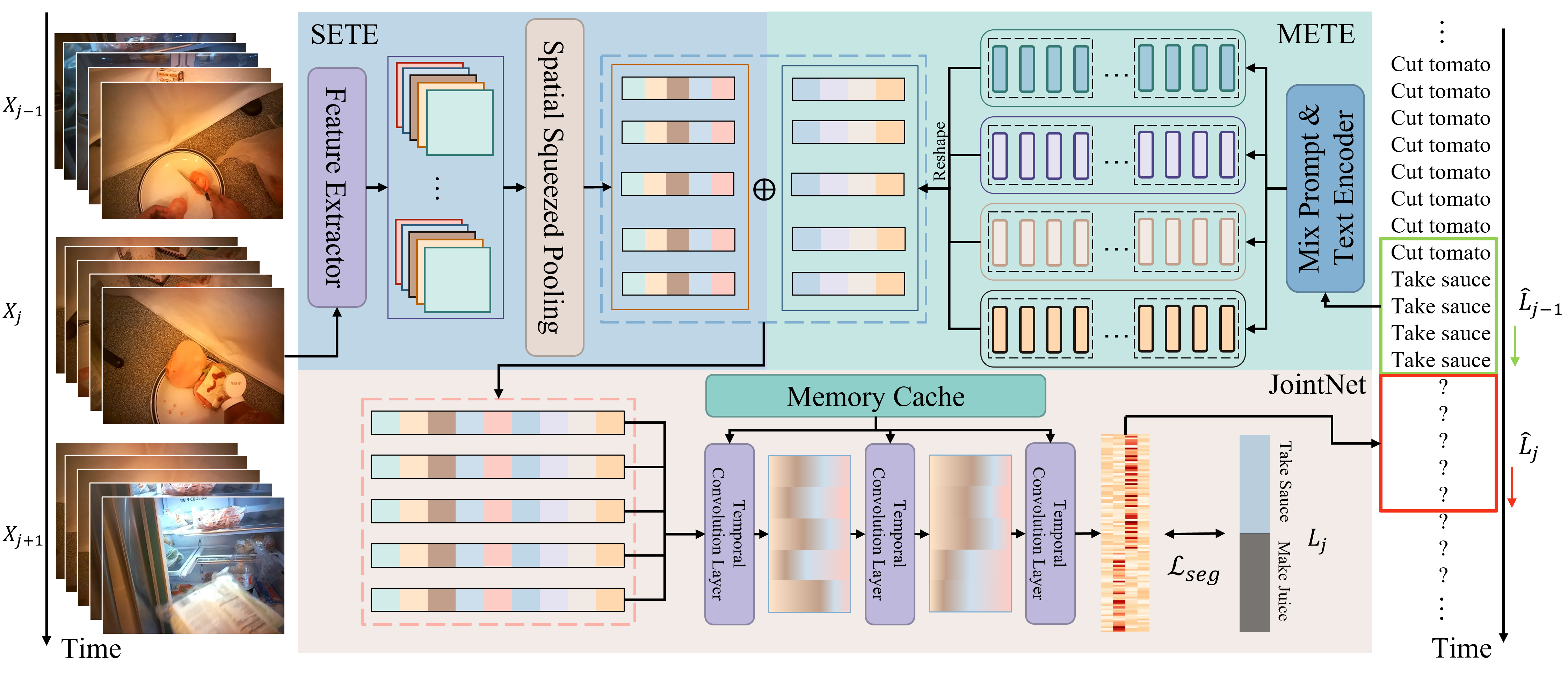}
\caption{Transeger model. This figure represents the training of Transeger model with $k = 5$. $\oplus$ represents the concatenation operation.}
\label{fig6}
\end{figure*}

\subsection{Transeger}
SETE and METE are just constructed to segment the current window of streaming video. They only model the current video sequence, ignoring information that has been seen before. We think SVTAS is very similar to  automatic speech recognition (ASR) in the processing of time sequences. Inspired by \cite{he2019streaming}, we make changes in our METE and propose Transeger, a  more reliable model for SVTAS task. See Figure \ref{fig6}. Transeger applies the idea of local connection of convolution networks to model the textual information of the previous video sequence. When Transeger infers each frame in the current moment,  it will combine previous textual information and current image information. More specifically, Transeger uses JointNet to blend text features and image features to jointly segment the action sequence of the current input streaming video sequence. Transeger consists of mix prompt engineering, text encoder, feature extractor, spatial squeezed pooling, and JointNet. They expect JointNet has been described in the previous section, so we will not repeat them here.

\subsubsection{JointNet}
In this paper, we use memory tcn to construct JointNet. JointNet first inverts the text features of the previous sequence. Then it contacts them with the image features of the current segment by feature dimension. Finally, the mixed features will be fed into the temporal model to segment current action sequence. JointNet is expressed in the following equations: step1: $F_{j}^{T,I} = downfall(F_{j-1}^{T}) \oplus F_{j}^{I_p}$, step2: $\widehat{L}_{j} = s(F_j^{T,I}; \theta_2)$. Where $\oplus$ denotes the concatenation operator, $F_{j-1}^{T} \in \mathbb{R}^{k \times d_t}$, $F_{j}^{I_p} \in \mathbb{R}^{k \times d_i}$ and $F_{j}^{T,I} \in \mathbb{R}^{k \times (d_t + d_i)}$, and $downfall(\cdot)$ represents the operator that inverts the matrix by the temporal dimension.

\subsubsection{Train}
For training, just like long short-term memory networks, teacher supervision is used to accelerate the training process. Specifically, $Pr_{j-1}$ is generated by $L_{j-1}$. After combining the text feature and image feature, we can use $\mathcal{L}_{seg}(\cdot, \cdot)$ as the loss function.

\subsubsection{Infer}
When inferring, the result of the previous video sequence inferred by model is used as the input of the current video sequence for mix prompt engineering. $Pr_{j-1}$ is generated by $\widehat{L}_{j-1}$.  We collect all segment results for evaluation. If the model applies in the full video, our model could use postprocessing \cite{li2021efficient} to improve the performance of discrimination of action completeness.

\section{Experiment}

\subsection{Datasets and Evaluation Metrics}
\subsubsection{Datasets}
The \textbf{GTEA} \cite{fathi2011learning} dataset contains 28 videos corresponding to 7 different activities, such as preparing coffee or cheese sandwiches, performed by 4 subjects. On average, there are 20 action instances per video. The evaluation was performed by excluding one subject to use cross-validation. The \textbf{50Salads} \cite{stein2013combining} dataset contains 50 videos with 17 action classes. On average, each video contains 20 action instances, and the average video length is 6.4 minutes. The activities were performed by 25 actors, each preparing two different salads. 50Salads also uses five-fold cross-validation. The \textbf{EGTEA} \cite{li2018eye} dataset is the largest of the three datasets. In total, EGTEA contains 28 hours (de-identified) of cooking activities from 86 unique sessions of 32 subjects. It contains 20 different actions, and each video contains an average of 15 action instances. Also, it will be evaluated by three-fold cross-validation.

\subsubsection{Evaluation Metrics}
To evaluate SVTAS task results, we adopt several metrics including frame-wise accuracy (\textbf{Acc}) \cite{farha2019ms}, mean average precision (mAP) metric with temporal IoU of 0.5 (denote by \textbf{mAP@0.5}) \cite{wang2022rcl}, the area under the AR (under specified temporal IoU thresholds for [0.5:0.05:1.0]) vs. AN (limiting the average number of proposals for each video and set to 100) curve (\textbf{AUC}) \cite{alwassel2021tsp} and the F1 score at temporal IoU threshold 0.1 (denote by \textbf{F1@0.1}) \cite{li2022bridge}. The frame-wise accuracy is the most direct and frequently used metric. mAP can evaluate the precision of model to detect action. AUC can measure the quality of the action segment proposal. F1 score is proposed to handle over-segmentation errors and measure the segmentation quality for action completeness.

\subsubsection{Implementation Details}
We adopt Adam optimizer with the base learning rate of $5 \times 10^{-4}$ with a $1 \times 10^{-4}$ weight decay. The spatial resolution of the input video is $256 \times 256$. We use Kinetics-400 \cite{carreira2017quo} pre-trained weight for all feature extractors. Each length of streaming video ($k$) is set to 32, with a sample rate 4 for GTEA and EGTEA, 20 for 50Salads. The model is trained 50 epochs on GTEA and 50Salads, and 35 epochs on EGTEA. The batch size is set to 2 during training.

\subsection{Results on Datasets}
\begin{table*}[!t]
\centering
\caption{Results on datasets. $k$ is set to 32. FPF means Flops Per Frame. FPMF means Flops Per Model  Forward. FPS means Frames Per Second. No means it can't be measured because this model is not real-time. We used RTX3090 platform to measure. $^{*}$ refers that the scores are reproduced by us. I3D+m-GRU+GTRM refers to the \cite{huang2020improving} model.}
\scalebox{0.95}{
    \begin{tabular}{@{}ccccccccccc@{}}
    \toprule
    Dataset                   & Training Stage                   & Model                           & Param(M)        & FPF(G)          & FPMF(G)          & FPS(Hz)       & Acc            & AUC            & mAP@0.5        & F1@0.1         \\ \midrule
    \multirow{5}{*}{GTEA}     & \multirow{2}{*}{Multi}  & ViT+asformer   & 54.645          & no              & \textbf{85.8432} & no            & \textbf{81}    & -              & -              & \textbf{94.1}  \\
                              &                         & I3D+ms-tcn     & \textbf{28.006} & no              & 173.12           & no            & 79.2           & 82.92$^{*}$    & 64.45$^{*}$    & 87.5           \\ \cmidrule(l){2-11} 
                              & \multirow{3}{*}{Single} & I3D+ms-tcn  & 28.006          & 5.41            & 173.12           & 433           & 43.93          & 59.2           & 13.03          & 42.44          \\
                              &                         & TSM+memory tcn+SETE(ours)       & \textbf{2.372}  & \textbf{0.3191} & \textbf{10.017}  & \textbf{3122} & 72.31          & \textbf{75.11} & \textbf{49.5}  & 70.29          \\
                              &                         & Transeger(ours)                 & 3.444           & 0.7955          & 25.457           & 1199          & \textbf{72.66} & 72.24          & 39.77          & \textbf{72.36} \\ \midrule
    \multirow{5}{*}{50Salads} & \multirow{2}{*}{Multi}  & ViT+asformer   & 54.645          & no              & \textbf{85.8432} & no            & \textbf{88.1}  & -              & -              & 89.2           \\
                              &                         & I3D+ms-tcn     & \textbf{28.006} & no              & 173.12           & no            & 80.7           & -              & -              & 76.3           \\ \cmidrule(l){2-11} 
                              & \multirow{3}{*}{Single} & I3D+ms-tcn        & 28.01           & 5.3995          & 172.786          & 433           & 30.31          & 48             & 7.13           & 12.36          \\
                              &                         & TSM+memory tcn+SETE(ours)       & \textbf{2.373}  & \textbf{0.313}  & \textbf{10.017}  & \textbf{3122} & 79.85          & 75             & 58.73          & 48.51          \\
                              &                         & Transeger(ours)                 & 3.477           & 0.7955          & 25.457           & 1199          & \textbf{82.51} & \textbf{75.68} & \textbf{59.22} & \textbf{54.99} \\ \midrule
    \multirow{5}{*}{EGTEA}    & \multirow{2}{*}{Multi}  & I3D+m-GRU+GTRM & -               & no              & -                & no            & \textbf{69.5}  & -              & -              & \textbf{41.6}  \\
                              &                         & I3D+ms-tcn     & 28.006          & no              & 173.12           & no            & 69.2           & -              & -              & 32.1           \\ \cmidrule(l){2-11} 
                              & \multirow{3}{*}{Single} & I3D+ms-tcn        & 28.011          & 5.41            & 173.12           & 433           & 59.43          & 2.94           & 0.6            & 2.76           \\
                              &                         & TSM+memory tcn+SETE(ours)       & \textbf{2.373}  & \textbf{0.3191} & \textbf{10.213}  & \textbf{3122} & \textbf{63.19} & \textbf{8.42}  & \textbf{3.27}  & \textbf{12.26} \\
                              &                         & Transeger(ours)                 & 3.482           & 0.7955          & 25.457           & 1199          & 60.89          & 7.01           & 1.9            & 11.52          \\ \bottomrule
    \end{tabular}
    }
	\label{tab2}
\end{table*}

\subsubsection{Evaluation Metrics}
The results of our model on three challenging datasets are shown in Table \ref{tab2}. Our best model achieves 90\% approximation in the relatively important metric Acc compared with the full-video state-of-the-art model. It is worth noting that we use less than 40\% of model computation compared with the full-video state-of-the-art model. And compared with the direct application of TAS model in real time, our models tower above its performance.

\subsubsection{Qualitative Results}
\begin{figure}[htbp]
    \centering
    \includegraphics[width=\linewidth]{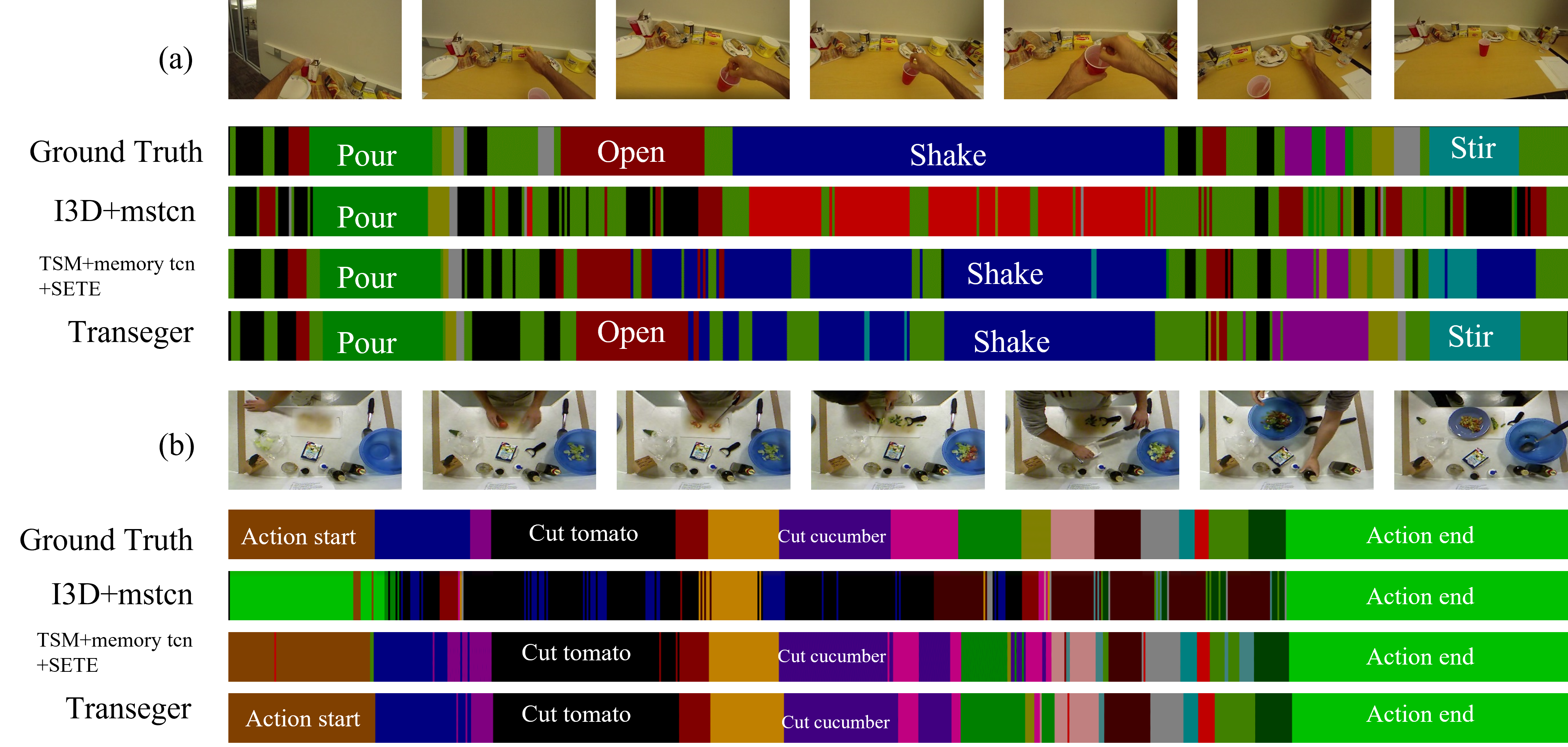}
    \caption{Comparison results of models on 50Salads and GTEA datasets. (a)  indicates GTEA dataset. (b) indicates 50Salads dataset.}
    \label{fig7}
\end{figure}

Figure \ref{fig7} shows the visual results of our model compared with other models on the GTEA and 50Salads dataset.

\subsubsection{Video Sample for Real-time Inferring}
\begin{figure}[htbp]
\centering
\includegraphics[width=\linewidth]{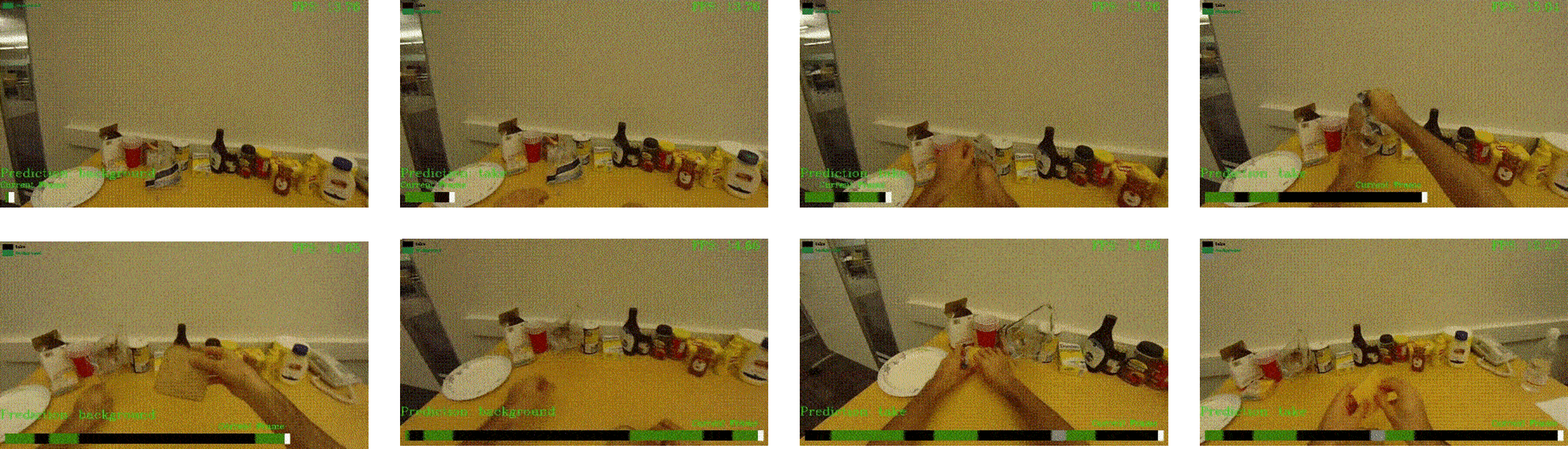}
\caption{Inference of GTEA datasets on the Intel 11th i7 CPU platform.}
\label{video_sample}
\end{figure}

We can observe that our model implements real-time inference, in Fighre \ref{video_sample}.

\begin{figure}[t]
\centering
\includegraphics[width=\linewidth]{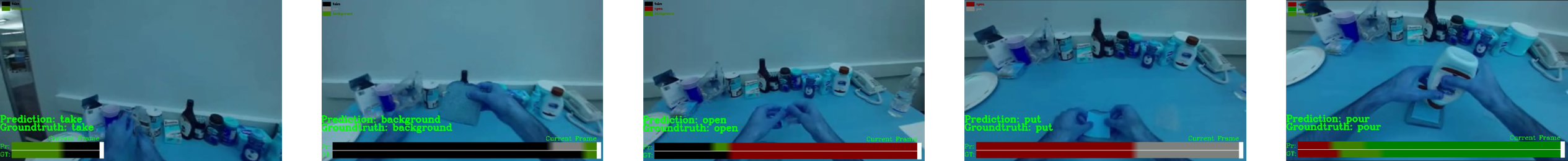}
\caption{Comparison of real-time inference results with ground truth.}
\label{pr_vs_gt}
\end{figure}

\subsection{Compare with models of related tasks}

\begin{table*}[!t]
 \centering
 \caption{Performance of the replicated model on the split1 of the GTEA dataset. RGB refers to raw image information of video. Flow means optical flow extracted from video. $^+$ is denoted as the average score of the cross-validation of GTEA, and the model score in split1 is usually lower than the average score of full validation, which indicates that the scores of $^+$ in the table are inflated. IC refers to image classification. VP refers to video prediction.}
    \scalebox{0.85}{
	\begin{tabular}{@{}l l c|c c c c|c c c c @{}} 
     \toprule[1pt]
     Publish   & Model                                                     & Task   & Real-Time & Train   Method & Modality         & $k$ Frames    & Acc   & AUC   & mAP@0.5 & F1@0.1 \\
     \midrule
     CVPR 2016 & I3D+bilstm\cite{singh2016multi}          & AR+TAS & no        & two stages  & RGB+flow         & full video & 73.09 & 74.58 & \textbf{65.43} & 83.51 \\
     CVPR 2019 & I3D+ms-tcn\cite{farha2019ms}             & AR+TAS & no        & two stages  & RGB+flow         & full video & 74.99 & \textbf{82.92} & 61.25 & 78.41 \\
     BMCV 2021 & I3D+asformer\cite{yi2021asformer}        & AR+TAS & no        & two stages  & RGB+flow         & full video & 75.33 & 75.63 & 56.86 & 83.98 \\
     CVPR 2022 & ViT+asformer\cite{li2022bridge}          & AR+TAS & no        & two stages  & RGB+flow         & full video & \textbf{81.00}$^+$ & - & - & \textbf{94.10}$^+$ \\
     \midrule
     -         & full connection layer                    & TAS    & yes      & two stages  & RGB+flow feature & 32       & 61.36 & 50.03 & 19.86 & 20.86 \\
     CVPR 2016 & bilstm\cite{singh2016multi}              & TAS    & yes      & two stages  & RGB+flow feature & 32       & 65.85 & 64.46 & 34.74 & 48.32 \\
     CVPR 2019 & ms-tcn\cite{farha2019ms}                 & TAS    & yes      & two stages  & RGB+flow feature & 32       & 51.08 & 55.25 & 31.19 & 30.04 \\
     BMCV 2021 & asformer\cite{yi2021asformer}            & TAS    & yes      & two stages  & RGB+flow feature & 32       & \textbf{69.80} & \textbf{73.32} & \textbf{42.56} & \textbf{73.45} \\
     ICCV 2021 & OadTR\cite{wang2021oadtr}         & OAD    & yes       & two stages      & RGB+flow feature & 32       & 67.76 & 67.03 & 41.01 & 53.69 \\
     \midrule
     CVPR 2016 & ResNet\cite{he2016deep}                  & IC     & yes       & single stage   & RGB              & 32        & 38.74 & 55.36 & 13.19 & 25.80 \\
     CVPR 2018 & MobileNetV2\cite{sandler2018mobilenetv2} & IC     & yes       & single stage   & RGB              & 32        & 40.94 & 51.45 &  9.18 & 30.54 \\
     ICLR 2021 & ViT\cite{dosovitskiy2020image}           & IC     & yes       & single stage   & RGB              & 32        & 28.70 & 29.31 &  1.67 & 21.88 \\
     ICLR 2022 & MobileViT\cite{mehta2021mobilevit}       & IC     & yes       & single stage   & RGB              & 32        & 25.66 & 21.36 &  1.85 & 17.91 \\
     CVPR 2017 & I3D\cite{carreira2017quo}                & AR     & yes       & single stage   & RGB              & 32        & 13.35 & 42.59 &  1.92 & 25.24 \\
     CVPR 2018 & R(2+1)D\cite{tran2018closer}            & AR     & yes       & single stage   & RGB              & 32        & 24.40 & 49.01 &  3.07 & 30.03 \\
     ICCV 2019 & TSM\cite{lin2019tsm}                     & AR     & yes       & single stage   & RGB              & 32        & 31.75 & 31.75 &  2.07 & 21.07 \\
     ICML 2021 & \footnotesize{TimeSformer\cite{bertasius2021space}}     & AR     & yes       & single stage   & RGB & 32    & 36.42 & 47.40 &  6.17 & 29.28 \\
     PAMI 2022 & PredRNNV2\cite{wang2022predrnn}          & VP     & yes       & single stage   & RGB              & 32        & 22.67 & 26.41 &  1.29 & 18.75 \\
     CVPR 2022 & ViT+asformer\cite{li2022bridge}          & AR+TAS & yes       & single stage   & RGB              & 32        & 23.40 &  0.00 &  0.00 &  0.00 \\
     CVPR 2019 & I3D+ms-tcn\cite{farha2019ms}             & AR+TAS & yes       & single stage   & RGB              & 32        & 43.82 & 60.83 & 13.85 & 34.70  \\
     CVPR 2016 & I3D+bilstm\cite{singh2016multi}          & AR+TAS & yes       & single stage   & RGB              & 32        & 53.92 & 56.51 & 21.99 & 58.73 \\
     -         & TSM+ms-tcn                               & AR+TAS & yes       & single stage   & RGB              & 32        & 44.57 & 59.00 & 21.45 & 35.29 \\
     ours      & I3D+tcn+SETE                             & SVTAS  & yes       & single stage   & RGB              & 32        & 57.38 & 65.16 & 34.20 & 61.68 \\
     ours      & TSM+tcn+SETE                             & SVTAS  & yes       & single stage   & RGB              & 32        & 70.37 & 73.48 & 46.34 & 62.56  \\
     ours      & TSM+memory tcn+SETE                      & SVTAS  & yes       & single stage   & RGB              & 32      & \textbf{73.86} & \textbf{75.02} & 49.00 & 68.18  \\
     ours      & TSM+memory tcn+METE                      & SVTAS  & yes       & single stage   & RGB              & 32        & 71.93 & 72.87 & 49.82 & 62.67 \\
     ours      & Transeger                                & SVTAS  & yes       & single stage   & RGB              & 32        & 73.18 & 71.16 & \textbf{52.68} & \textbf{71.69}  \\
     \bottomrule
    \end{tabular}
	}
    \label{tab1}
\end{table*}

We evaluate the relevant models that may be applied to the SVTAS task on GTEA dataset. The results are shown in Table \ref{tab1}.

The experimental results show that our model is close to the full video TAS model in terms of Acc and AUC (Task AR+TAS vs. Task SVTAS), while other models for SVTAS task perform badly (Train Method of single stage). Our model has a large gap with the non-real-time TAS model in the measure of action completeness (F1@0.1) because model can only get information at the current moment. But in exchange, the model can perform real-time TAS and can be trained more effectively, which we consider acceptable. Moreover, by comparing the image classification model and the action recognition model, we can find that the image classification model slightly outperforms the action recognition model on the SVTAS task. It is because the image classification model is more suitable for the one-to-one paradigm.

Training the TAS model directly with streaming features also could get an acceptable result. We think it is because video feature is an effective abstract representation of video. But they are based on RGB modalities and optical flow modalities. Both need additional work for training. Our model gives better results using only RGB modalities and is end-to-end trained. Among our models, TSM+memory tcn+SETE performs best on Acc metrics; and the multi-modality model Transeger is optimal in the metrics of detecting action (mAP@0.5) and the metrics of measuring action completeness (F1@0.1). It is because Transeger model could get the information from the previous segment in streaming video. And this information could significantly reduce the false positive results of the model. 

Although the full-video model with two-stage training is currently high in all metrics, we point out the possibility of end-to-end training models and real-time TAS models in this paper.

\subsection{Compare with spatio-temporal modeling approaches}

\begin{table}[htbp]
	\centering
    \caption{Compare with different spatio-temporal modeling approaches.}
	\scalebox{0.7}{
	\begin{tabular}{@{}l c|c c c c@{}} 
     \toprule
     Modeling Approaches & Model              & Acc & AUC & mAP@0.5 & F1@0.1 \\
     \midrule
     3D                         & I3D+tcn+SETE       & 57.38 & 65.16 & 34.20 & 61.68 \\
     2D+1D                      & R(2+1)D+tcn+SETE   & 67.10 & 70.64 & 43.34 & \textbf{63.66} \\
     2D+temporal shift          & TSM+3Dtcn       & 69..35 & 68.47 & 40.19 & 62.65 \\
     2D+temporal shift          & TSM+tcn+SETE       & \textbf{70.37} & \textbf{73.48} & \textbf{46.34} & 62.56 \\
     \midrule
     3D                         & I3D+memory tcn+SETE & 64.77 & 67.47 & 38.29 & 64.37 \\
     2D+1D                      & R(2+1)D+memory tcn+SETE & 67.02 & 68.62 & 35.84 & \textbf{70.46} \\
     2D+temporal shift          & TSM+memory tcn+SETE & \textbf{73.86} & \textbf{75.02} & \textbf{49.00} & 68.18 \\
     \bottomrule
    \end{tabular}
	}
	\label{tab3}
\end{table}

We compared the different spatio-temporal modeling approaches of the feature extractor. The results are shown in Table \ref{tab3}. The experiments show that the combination of 2D convolution operator \cite{sandler2018mobilenetv2} and temporal shift module \cite{lin2019tsm} outperforms 2D plus 1D convolution operator \cite{tran2018closer} over 3D convolution operator \cite{carreira2017quo}. We conjecture that the combination of 2D convolution operator \cite{sandler2018mobilenetv2} and temporal shift module \cite{lin2019tsm} focuses on the spatial comparison between frames in the feature extraction phase and temporal aggregation in the segmentation phase. And, this pattern is more efficient than others. Model without spatial squeezed pooling takes up more VRAM instead of improving model performance (Line 4 vs. Line 5). The experimental results in Table \ref{tab3} also show that the temporal convolution network with memory cache (memory tcn) which we select, can improve the performance of real-time TAS based on various spatio-temporal modeling approaches. The past information can indeed improve the performance of  SVTAS task.

\subsection{Ablation Studies}
\subsubsection{Mix prompt ablation study}

\begin{table}[htbp]
    \centering
    \caption{Component of mix prompt ablation study on Transeger. Because action category information is necessary, we fail to conduct an ablation study for it.}
    \scalebox{0.75}{
    \begin{tabular}{@{}cccc|cccc@{}}
    \toprule
    \makecell[c]{Ordinal \\ Prompt} & \makecell[c]{Action \\ Duration \\ Prompt} & \makecell[c]{Relative \\ Position \\ Prompt} & \makecell[c]{Learnable \\ Action \\ Category \\ Prompt} & Acc & AUC & mAP@0.5 & F1@0.1 \\
    \midrule
    \Checkmark &            &            & \Checkmark & 71.65 & 68.06 & 47.77 & 68.33 \\
    \Checkmark & \Checkmark &            & \Checkmark & 73.04 & 71.05 & 50.27 & 64.91 \\
               & \Checkmark & \Checkmark & \Checkmark & \textbf{73.76} & \textbf{72.42} & \textbf{58.68} & 69.87 \\
    \Checkmark & \Checkmark & \Checkmark & \Checkmark & 73.18 & 71.16 & 52.68 & \textbf{71.69} \\
    \bottomrule
    \end{tabular}
    }
    \label{tab4}
\end{table}

Table \ref{tab4} shows the ablation experiments for the components of mix prompt. The performance of model improves significantly after adding the action duration prompt. This is because the information on action duration helps the model infer the completeness of the action. The combination of relative position prompt and action duration prompt can help the model identify action segments when actions are about time mirror. From the experiments, the ordinal prompt only improves the model's ability to discriminate actions completeness. We infer the reason is that we use relative ordinal prompt instead of absolute ordinal prompt. An action can be either the first or the last, which may confuse the model in classifying some frame-level actions.

\subsubsection{Hyper-parameters $k$}

\begin{table}[htbp]
	\centering
    \caption{The ablation study of $k$ on split1 of GTEA dataset.}
	\scalebox{0.9}{
	\begin{tabular}{@{}l c|c c c c c@{}} 
     \toprule
     $k$ & Model & VRAM(G) & Acc & AUC & mAP@0.5 & F1@0.1\\ 
     \midrule
     8  & Transeger  & 4.7  & 65.68 & 67.77 & 41.52 & 66.28 \\
     16 & Transeger  & 6.6  & 70.60 & 63.92 & 52.26 & 66.66 \\
     32 & Transeger  & 8.2  & \textbf{73.18} & \textbf{71.16} & \textbf{52.68} & \textbf{71.69} \\
     64 & Transeger  & 18.6 & 69.29 & 67.35 & 46.34 & 58.88 \\
     \bottomrule
    \end{tabular}
	}
	\label{tab5}
\end{table}

From Table \ref{tab5}, we can see that the model performs best when the segment length of the streaming video is 32. We think that it is related to the average length of the action. To enhance the performance of the model on certain datasets, the model should be adapted for datasets with different length distributions by adjusting the segment length $k$ or setting an appropriate sample rate.

\subsubsection{More experiments for compare with models of related tasks}
\begin{table}[hbtp]
 \centering
 \caption{Performance of model on the split1 of the 50salads dataset. Flow means optical flow extracted from video. IC refers to image classification. VP refers to video prediction.}
    \scalebox{0.45}{
	\begin{tabular}{@{}l l c|c c c c|c c c c @{}} 
     \toprule
     Publish   & Model                                                     & Task   & Real-Time & Train   Method & Modality         & $k$ Frames    & Acc   & AUC   & mAP@0.5 & F1@0.1 \\
     \midrule
     CVPR 2019 & I3D+ms-tcn\cite{farha2019ms}             & AR+TAS & no        & two stages  & RGB+flow         & full video & 80.70 & - & - & 76.30 \\
     BMCV 2021 & I3D+asformer\cite{yi2021asformer}        & AR+TAS & no        & two stages  & RGB+flow         & full video & \textbf{85.60} & - & - & \textbf{85.10} \\
     \midrule
     CVPR 2019 & ms-tcn\cite{farha2019ms}                 & TAS    & yes      & two stages  & RGB+flow feature & 32       & 11.64 & 41.90 & 2.60 & 4.99 \\
     \midrule
     CVPR 2017 & I3D\cite{carreira2017quo}                & AR     & yes       & single stage   & RGB              & 32        & 12.65 & 48.99 &  2.32 & 6.6 \\
     CVPR 2019 & I3D+ms-tcn\cite{farha2019ms}             & AR+TAS & yes       & single stage   & RGB              & 32        & 32.25 & 49.86 & 10.33 & 12.39 \\
     ours      & TSM+memory tcn+METE                      & SVTAS  & yes       & single stage   & RGB              & 32        & 77.92 & 71.51 & 54.94 & 45.88 \\
     ours      & Transeger                                & SVTAS  & yes       & single stage   & RGB              & 32        & \textbf{80.65} & \textbf{72.78} & \textbf{56.09} & \textbf{56.50}  \\
     \bottomrule
    \end{tabular}
	}
    \label{more_ex}
\end{table}

From the Table.\ref{more_ex}, we can see that our model is more suitable for TAS in real-time scenes.

\section{Conclusion}
In this paper, we focus on the cost of model training and the applicable limitation of TAS model. Firstly, for reducing the cost of model training, we propose three different end-to-end training frameworks to get rid of complicated step -- extracting frame-level features. Secondly, we propose Transeger model to make multi-modality real-time TAS realize. Finally, our model broadens the application scenarios of the TAS. SVTAS model can be applied for offline, online, streaming, and real-time scenarios, and so on. The widespread experiment results show that our proposed multi-modality real-time SVTAS model can use less than 40\% of state-of-the-art model computation to achieve 90\% of full-video state-of-the-art performance on challenging datasets. We should point out that although training on large datasets is attractive, we are constrained by computational resources.

\bibliographystyle{IEEEtran}
\bibliography{refer}
\end{document}